\documentclass[sigconf]{acmart}
\usepackage{subcaption}
\usepackage{multirow}
\usepackage{enumitem}
\AtBeginDocument{%
  }

\setcopyright{acmlicensed}
\copyrightyear{2018}
\acmYear{2018}
\acmDOI{XXXXXXX.XXXXXXX}
\acmConference[Conference acronym 'XX]{Make sure to enter the correct
  conference title from your rights confirmation email}{June 03--05,
  2018}{Woodstock, NY}
\acmISBN{978-1-4503-XXXX-X/2018/06}

\begin{document}

\title{PEARL: Unbiased Percentile Estimation via Contrastive Learning for Industrial-Scale Livestream Recommendation}

\author{Blake Gella}
\authornote{Core contributors.}
\email{blake.gella@tiktok.com}
\affiliation{%
  \institution{TikTok}
  \city{San Jose}
  \state{California}
  \country{USA}
}
\author{Wei Wu}
\authornotemark[1]
\email{wuwei.bupt@bytedance.com}
\affiliation{%
  \institution{ByteDance}
  \city{San Jose}
  \state{California}
  \country{USA}
}

\author{Yuhao Yin}
\authornotemark[1]
\email{yuhao.yin@tiktok.com}
\affiliation{%
  \institution{TikTok}
  \city{San Jose}
  \state{California}
  \country{USA}
}

\author{Zexi Huang}
\authornotemark[1]
\authornote{Corresponding author.}
\email{zexi.huang@tiktok.com}
\affiliation{%
  \institution{TikTok}
  \city{San Jose}
  \state{California}
  \country{USA}
}

\author{Zikai Wang}
\email{wangzikai.kevin@bytedance.com}
\affiliation{%
  \institution{ByteDance}
  \city{San Jose}
  \state{California}
  \country{USA}
}

\author{Emily Liu}
\email{emily.zfliu@bytedance.com}
\affiliation{%
  \institution{ByteDance}
  \city{San Jose}
  \state{California}
  \country{USA}
}

\author{Junlin Zhang}
\email{zhangjunlin.neicul@bytedance.com}
\affiliation{%
  \institution{ByteDance}
  \city{San Jose}
  \state{California}
  \country{USA}
}

\author{Wentao Guo}
\email{wentao.guo@bytedance.com}
\affiliation{%
  \institution{ByteDance}
  \city{San Jose}
  \state{California}
  \country{USA}
}

\author{Qinglei Wang}
\email{wangqinglei@bytedance.com}
\affiliation{%
  \institution{ByteDance}
  \country{Singapore}
}

\renewcommand{\shortauthors}{Gella$^*$, Wu$^*$, Yin$^*$, Huang$^*$, et al.}

\begin{abstract}
Recommender systems trained on user interaction data are susceptible to behavioral intensity imbalance---a systematic distortion arising from heterogeneous engagement patterns across users. This imbalance skews feedback signals such that observed interactions no longer faithfully reflect true preferences, causing models to disproportionately amplify signals from highly active users while underrepresenting others, which ultimately degrades recommendation quality and robustness at scale. To address this issue, we propose a nonparametric contrastive percentile approximation framework, PEARL, that models relative preference signals instead of absolute engagement magnitudes. Building upon relative advantage debiasing, PEARL leverages real contrastive interaction samples to approximate percentile relationships directly, without relying on auxiliary distribution estimation models. We provide theoretical justification demonstrating that such pairwise comparisons yield unbiased estimates of percentile-based preference signals. For broader applicability, we introduce a prediction-based bootstrapping mechanism for percentile smoothing to handle sparse and discrete feedback, alongside a generalized value-weighted formulation and a co-training strategy to enhance both modeling flexibility and representation learning. Extensive offline experiments demonstrate that PEARL effectively mitigates behavioral bias and consistently improves recommendation performance across multiple ranking targets. Deployed in a production livestream platform with a combined user base of billions, online A/B testing confirms substantial real-world gains: +2.10\% Watch Duration, +0.80\% Consumption Amount, +1.49\% Interaction Rate, and -6.91\% Report Rate.

\end{abstract}








\maketitle

\section{Introduction}

Modern recommender systems rely on large-scale user interaction data to learn personalized models that optimize engagement, satisfaction, and platform value ~\cite{zhang2024wukong}. Central to this paradigm is the assumption that observed feedback---such as clicks, watch time, or consumption-based interactions---serves as a reliable proxy for underlying user preference. In practice, however, user interactions are shaped not only by intrinsic interests but also by systematic behavioral heterogeneity: some users generate dense signals through frequent activity or high-value actions, while others interact sparsely or passively ~\cite{li2017neural,zhou2018deep}. Such disparities in behavioral intensity produce highly uneven feedback distributions, where a small subset of highly active or high-spending users disproportionately dominates the training signal ~\cite{parknalexander2008, himan2019}. This imbalance creates a structural mismatch between observed interactions and genuine preference, biasing learned models toward patterns driven by activity level or spending propensity rather than true user intent ~\cite{joachims2017unbiased,wang2018deconfounded}.

The mismatch between observed interactions and true preference has motivated extensive research on debiasing techniques in recommender systems. A prominent line of work addresses bias through counterfactual learning and propensity-based reweighting, adjusting interaction samples to better approximate unbiased preference signals ~\cite{swaminathan2015self,schnabel2016recommendations}. Another family of approaches adopts causal inference frameworks to model the relationship between user behavior and observed feedback, aiming to disentangle intrinsic interest from confounding effects ~\cite{bonner2018causal,zhan2022deconfounding}. More recently, representation-level debiasing methods have been proposed to prevent deep models from encoding spurious behavioral artifacts directly into user embeddings ~\cite{liang2018variational,ma2019learning,chen2021autodebias}. Despite their effectiveness, these approaches largely treat bias as a property of the interaction data distribution while implicitly assuming that user feedback follows homogeneous behavioral patterns.

A more direct avenue for addressing engagement bias has emerged through relative preference modeling. Rather than correcting raw feedback, RAD ~\cite{rad2025} proposes transforming watch-time observations into relative advantage signals by estimating conditional engagement distributions and converting them into percentile-based preference scores, thereby normalizing heterogeneous engagement patterns across users. While effective, this approach depends on an auxiliary model to approximate engagement distributions, introducing additional modeling complexity and potential estimation error. Motivated by this perspective, we propose PEARL, a lightweight yet principled framework that instead approximates relative preference signals through contrastive samples observed naturally in training data. For each user, PEARL constructs pairwise comparison signals between interactions, learning to predict relative preference orderings from observed engagement via a binary cross-entropy objective. By leveraging such naturally occurring comparisons, PEARL eliminates the need for explicit distribution estimation, offering a simpler and nonparametric path to relative advantage modeling while directly mitigating behavioral intensity bias.

The main contributions of this paper are as follows:
\begin{itemize}[leftmargin=*]
\item \textbf{User debiasing framework}: We identify behavioral intensity imbalance as a key source of user-side bias in recommender systems, where heterogeneous engagement behaviors lead to skewed interaction signals and distorted preference learning. To address this, we propose a nonparametric contrastive percentile approximation framework that derives relative preference signals from real interaction comparisons, avoiding explicit distribution estimation. We further provide theoretical justification showing that such contrastive comparisons constitute an unbiased approximation of percentile-based relative advantage signals.

\item \textbf{Extensions for real-world recommendation}: Building upon the core formulation, we develop several practical extensions to improve robustness and applicability: a prediction-based bootstrapping mechanism for percentile smoothing, which alleviates label sparsity by replacing coarse empirical distributions with continuous prior predictions; a value-weighted formulation that accounts for heterogeneous interaction importance; and a co-training strategy that enriches representation learning through complementary supervision signals. Together, these extensions enable PEARL to handle sparse, discrete feedback and adapt to diverse recommendation scenarios.

\item \textbf{Empirical effectiveness}: Extensive offline experiments demonstrate that PEARL consistently mitigates behavioral bias and improves recommendation performance in terms of User AUC across multiple ranking targets. Further, a series of online A/B tests show that PEARL achieves +2.1\% Watch Duration, +0.80\% Consumption Amount, +1.49\% Interaction Rate, and -6.91\% Report Rate in our production recommender system, a leading online livestreaming platform that attracts billions of users globally.

\end{itemize}

\section{Related Work}

\subsection{User Action Modeling}
Modern recommendation systems have been developed to deliver personalized content through user engagement. This engagement can be quantified by various user actions, like watch time, longterm retention, and consumption-based interactions. Out of these actions, regression targets, such as watch time and consumption amount, have been a focus of many previous works, mainly for high correlation for user interest and engagement and their difficulty in modeling. Early techniques for watch time, like Weighted Logistic Regression ~\cite{covington2016deep}, use watch times to weight the impression in training. CREAD ~\cite{sun2024cread} introduces training optimizations through discretization, classification, and restoration to address the long-tailed distribution of watch time. These methods optimize model training, but do not account for the inherent biases in the features of an instance, such as duration bias or exposure bias. As such, these methods negatively impact user bias, as samples with high values get overrepresented in training.

\subsection{Capturing Modeling Biases}
To address the biases from different sources, various works have proposed solutions for debiasing the target training. DVR ~\cite{zheng2022dvr} uses adversarial learning to specifically remove the bias of video duration. D2Co ~\cite{zhao2023uncovering} targets both the duration and noise biases, using a Gaussian Mixture Model to estimate both. However, with long-tailed distributions, these methods can lead to poor generalization due to data distribution. To correct for this, researchers propose quantile regression strategies to better capture the full distribution of watch time rather than a single expected value. For instance, D2Q ~\cite{zhan2022deconfounding} segments videos by duration and fits quantile regression within each group, while CQE ~\cite{lin2024conditional} extends this by modeling the complete watch-time distribution, enabling more flexible inference under skewed or uncertain engagement patterns. RAD ~\cite{rad2025} builds on these works by correcting watch time relative to empirically derived reference distributions conditioned on both users and items, addressing multiple biases related to the user, item, and recommendation system. While these methods work effectively in the setting of video recommendation, they do not efficiently transfer to the context of livestream recommendation.

\subsection{Debiasing in Livestream Recommendation}

Livestream recommendation presents unique challenges related to biases due to the characteristics of its content format. Unlike video recommendation, livestreaming is not heavily biased by item duration, as the average watch time is much smaller than the average livestream's duration. Other works have made efforts to optimize livestream recommendation modeling by incorporating more comprehensive signals, like video and audio in ContentCTR ~\cite{deng2023contentctr}, or training configuration optimization, like the multiple time windows proposed by ~\cite{zhao2025towards}. On mixed-format platforms, users' strong preference toward a particular content format, such as short-form video, can introduce bias, where alternative formats are skipped despite potentially aligning with the user's underlying interests. Thus, a user's behavior with livestreaming content may differ drastically compared to their short video behavior, leading to larger variances in user behaviors in livestream recommendation. We propose a debiasing method that disentangles content interest from format disinterest in livestream recommendation, recovering meaningful signals that engage users with limited interaction signals. Our method to capture user interest for the livestream format from sparse interaction signals, with applications for other minority-supply content in an app beyond livestreams. Compared to previous works, our method is centered on the challenges faced by livestream recommendation in a mixed format setting, and, thus, we focus on the issue of user bias instead of duration bias.
\section{Method}

Directly regressing on continuous percentile values within the $[0, 1]$ interval is non-trivial in industrial-scale recommender systems. This complexity arises from two primary hurdles: the systematic overhead of real-time distribution tracking and the intrinsic modeling challenges associated with percentile regression. More fundamentally, as motivated in Section~\ref{sec:motivation}, a robust ranking framework must effectively disentangle user-side behavioral intensity from intrinsic preference to resolve the inherent user bias.

\begin{figure*}[t]
    \centering
    \includegraphics[width=0.95\textwidth]{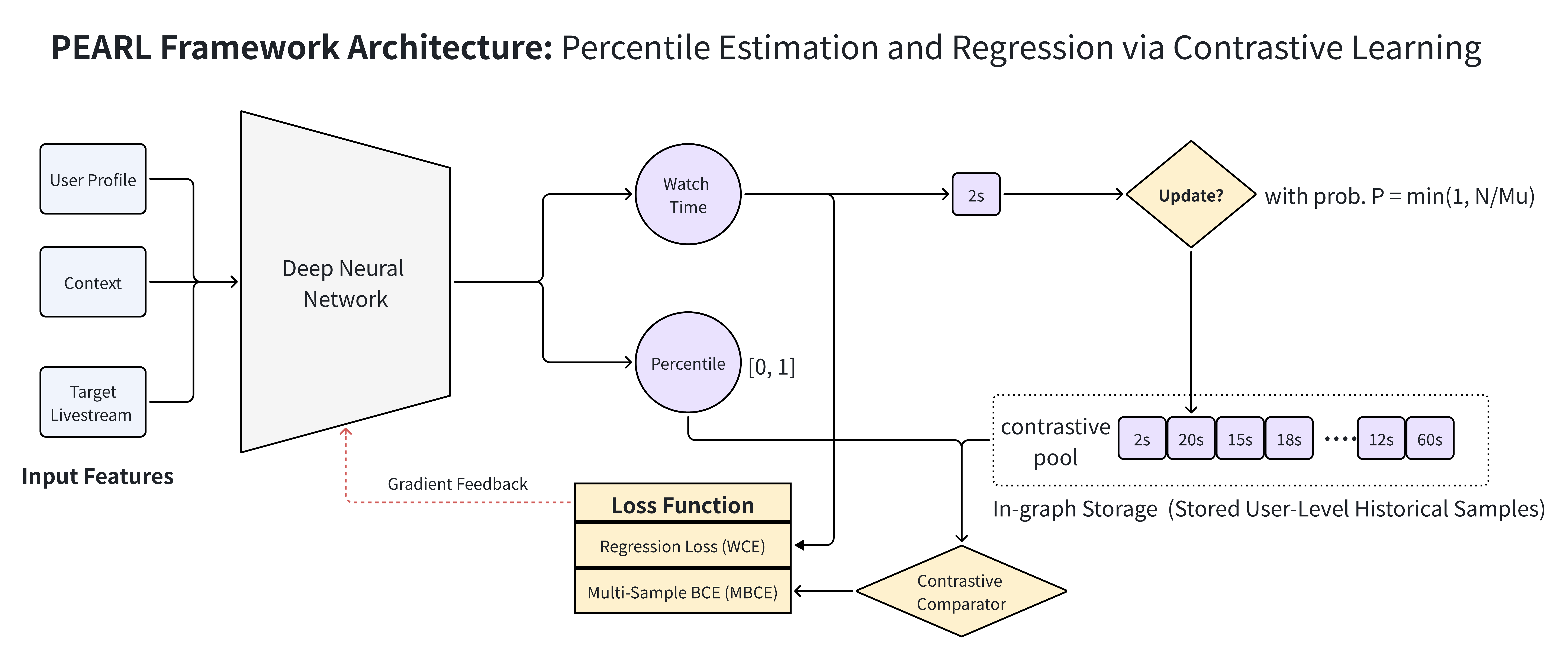} 
    \caption{The overall architecture of PEARL, which consists of two core components: 
    A user-keyed reservoir pool ($N = 50$) that dynamically maintains an unbiased historical distribution. (2) Contrastive Learning: A multi-sample comparator that cross-evaluates model predictions against historical samples to compute the multi-sample contrastive loss ($\mathcal{L}_{MBCE}$) without explicit distribution tracking.}
    \label{fig:pearl_framework}
\end{figure*}

To address these challenges, we introduce our \underline{P}ercentile \underline{E}stimation \underline{a}nd \underline{R}egression via Contrastive \underline{L}earning (\textbf{PEARL}) framework for recommender systems. PEARL shifts the prediction target from raw magnitudes (e.g., watch time) to relative user-wise percentiles using a distribution-free contrastive learning paradigm, thereby recovering consistent preference signals across heterogeneous user groups. An overview of the PEARL framework is systematically illustrated in Fig.~\ref{fig:pearl_framework}.

This section is organized as follows. We first elaborate on the limitations of standard regression modeling and the motivation for percentile modeling (Section~\ref{sec:motivation}). We then derive an unbiased percentile estimator via \textbf{single-sample contrastive learning} (Section~\ref{sec:Single-Sample}) and introduce \textbf{multi-sample contrastive learning} to reduce estimation variance (Section~\ref{sec:Multi-Sample}). To bridge the gap between relative ranking and reward magnitude, we extend the framework to a generalized \textbf{value-weighted contrastive learning} (Section~\ref{sec:Value-Weighted}). Furthermore, we address the challenge of discrete classification targets (e.g., clicks or gifts) via \textbf{prediction-based bootstrapping} (Section~\ref{sec:Bootstrapped Contrastive}) and propose a \textbf{co-training strategy} to enhance the ranking capability of standard regression targets (Section~\ref{sec:Multi-Task Contrastive}). Finally, we cover the practical considerations in the implementation of PEARL in billion-scale recommender systems (Section~\ref{sec:Billion-Scale System}).

\subsection{Problem Formulation and Motivation}
\label{sec:motivation}
In recommender systems, ranking models are typically designed to predict user preferences among multiple candidate items for a given user. Standard regression paradigms often attempt to fit continuous implicit or explicit feedback targets, such as watch time (e.g., the duration of user engagement) or interaction counts (e.g., the frequency of comments or gifts). However, these approaches focus primarily on fitting the magnitude of the targets while overlooking their relative ranks. Within a ranking context, these relative ranks are more indicative of a user's actual decision-making process, as preferences are inherently revealed through comparison rather than the magnitude.

When facing a diverse user population with heterogeneous behaviors, regression on raw magnitudes introduces significant user bias. A fixed target magnitude represents disparate levels of interest for users with varying behavioral intensities. For example, a 30-second watch time might only indicate moderate interest (e.g., 50th percentile) for a highly active user who consumes livestream content daily. Conversely, for a low-activity user, the same 30 seconds could signify exceptionally high interest (e.g., 99th percentile).

 Rather than attempting to eliminate this inherent bias, we incorporate user-specific behavioral context by shifting the prediction target toward relative percentile prediction. For an interaction with engagement magnitude $y \in \mathbb{R}^+$ (e.g., watch time) for user $u$, let $F_u(\cdot)$ denote the Cumulative Distribution Function (CDF) of user $u$'s historical engagement. We define the ground-truth percentile $p$ as:
\begin{equation}
    p = F_u(y) = P(Y_u \le y)
\end{equation}
where $Y_u$ is a random variable representing the engagement magnitudes of user $u$. 

\noindent\textbf{Formal Task Definition.} Given an interaction between user $u$ and item $i$, denote the input feature vector as $\mathbf{x} = [\mathbf{x}_u, \mathbf{x}_i, \mathbf{x}_c] \in \mathcal{X}$, where $\mathbf{x}_u, \mathbf{x}_i$, and $\mathbf{x}_c$ represent user profiles, item metadata, and contextual signals, respectively. Our objective is to learn a predictive mapping function $f: \mathcal{X} \rightarrow [0, 1]$, parameterized by $\theta$, such that the predicted score $\hat{p}$ approximates the ground-truth user-specific percentile:
\begin{equation}
    f(\mathbf{x}; \theta) = \hat{p} \approx F_u(y)
\end{equation}
The optimization goal is to find the optimal parameters $\theta^*$ that minimize the expected risk over the joint distribution of features and engagement magnitudes:
\begin{equation}
    \theta^* = \arg\min_{\theta} \mathbb{E}_{(\mathbf{x}, y) \sim \mathcal{D}} \left[ \ell(f(\mathbf{x}; \theta), F_u(y)) \right]
\end{equation}
where $\ell(\cdot)$ is a ranking-compatible loss function. By fitting the percentile $F_u(y)$ instead of the raw magnitude $y$, we transform heterogeneous behavioral signals into a standardized, bias-free preference space.

However, directly performing regression on percentile values within the $[0, 1]$ interval is highly non-trivial in industrial-scale streaming recommender systems, primarily due to two fundamental challenges:
\begin{itemize}[leftmargin=*]
    \item \textbf{Systematic Overhead of Distribution Tracking:} Real-time percentile acquisition is computationally and storage-prohibitive. Precisely capturing $F_u(y)$ requires maintaining the exact historical distribution for users across industrial-scale platforms (\textit{Storage Cost}) and performing real-time sorting operations during inference (\textit{Computational Latency}), which is infeasible under strict millisecond-level constraints.
    \item \textbf{Intrinsic Challenges in Direct Percentile Regression:} Even if ground-truth percentiles were available, directly fitting $p \in [0, 1]$ via standard regression objectives (e.g., Mean Squared Error or Weighted Cross Entropy) is inherently difficult. These losses often struggle with the bounded and non-linear nature of percentile distributions, where small estimation errors in high-percentile regions can lead to significant ranking suboptimality.
\end{itemize}

In the following sections, we introduce a novel and lightweight contrastive learning approach to overcome these challenges without direct regression modeling.

\subsection{Single-Sample Contrastive Learning}
\label{sec:Single-Sample}

To address the challenges of direct percentile regression, we introduce a Single-Sample Contrastive Learning paradigm. This approach elegantly transforms the continuous percentile regression task into a binary contrastive problem, requiring only a single stochastic sample from the user's history as a reference point.

\textbf{Contrastive Label Construction.} Suppose the engagement magnitude of the current training instance for user $u$ is $y$ (e.g., watch time). To estimate its percentile, we sample a historical target value $Y'$ from the same user's historical distribution $f_u(\cdot)$ as a contrastive reference. We then define a binary indicator function $\mathbb{I}(y > Y')$ as the contrastive target:
\begin{equation}
\mathbb{I}(y > Y') = 
\begin{cases} 
1, & \text{if } y > Y' \\ 
0, & \text{otherwise} 
\end{cases}
\end{equation}

We now demonstrate that optimizing toward this binary contrastive objective allows the model to capture the underlying percentile without explicit distribution assumptions.

\textbf{Theorem 1 (Unbiased Estimator of Percentile).} 
Let $Y'$ be a random variable representing the historical engagement magnitude of user $u$, following a probability density function $f_u(\cdot)$. For a training instance with a fixed magnitude $y$, the expectation of the binary indicator $\mathbb{I}(y > Y')$ is the Cumulative Distribution Function (CDF) value of $y$, representing the ground-truth percentile:
\begin{equation}
\mathbb{E}_{Y' \sim f_u} [\mathbb{I}(y > Y')] = \int_{0}^{y} f_u(t) dt = \text{CDF}_u(y)
\end{equation}

\textit{Proof:} 
By the fundamental property of indicator variables, the expectation is the probability of the event $\{Y' < y\}$. Since $y$ is a fixed realization for the current instance and $Y'$ is a random variable sampled from the user's historical distribution $f_u(\cdot)$, we have:
\begin{equation}
\mathbb{E}_{Y'} [\mathbb{I}(y > Y')] = 1 \cdot P(Y' < y) + 0 \cdot P(Y' \ge y) = P(Y' < y)
\end{equation}
By the definition of the CDF for a continuous random variable $Y'$, the probability $P(Y' < y)$ is given by the integral of its density function $f_u(\cdot)$ from its lower bound to $y$:
\begin{equation}
P(Y' < y) = \int_{0}^{y} f_u(t) dt = \text{CDF}_u(y)
\end{equation}
This confirms that the expectation of our contrastive signal is an unbiased estimator of the true percentile.

\textbf{Single-sample Binary Cross-Entropy Loss.} To optimize the model toward this unbiased target, we minimize the Binary Cross-Entropy (BCE) loss between the predicted percentile $\hat{p} = f(\mathbf{x}; \theta)$ and the contrastive indicator:
\begin{equation}
    \mathcal{L} = - [\mathbb{I}(y > Y') \log(\hat{p}) + (1 - \mathbb{I}(y > Y')) \log(1 - \hat{p})]
\end{equation}
Since the expectation of the contrastive label is the true percentile (as per Theorem 1), minimizing this objective ensures that the model's prediction $\hat{p}$ converges to the unbiased percentile $F_u(y)$.

In summary, while the single-sample approach establishes a theoretically sound and unbiased foundation for percentile estimation, the reliance on a single stochastic reference point introduces considerable training variance. This stochasticity can lead to noisy gradient updates in billion-scale streaming environments, necessitating a more robust multi-sample strategy to stabilize the learning process, which we explore in the following section.

\subsection{Multi-Sample Contrastive Learning}
\label{sec:Multi-Sample}
While the estimator based on a single historical contrastive sample is mathematically unbiased (as shown in Theorem 1), it suffers from high variance due to the stochastic nature of the sampling process, which introduces significant noise during model training. In this section, we quantify this variance and propose a multi-sample optimization strategy to stabilize the learning process.

\textbf{Theorem 2 (Quantitative Variance Analysis).} 
Consider the binary contrastive indicator $B = \mathbb{I}(y > Y')$ for a single pair. As a Bernoulli random variable with success probability $p = \text{CDF}_u(y)$, its variance is $\text{Var}(B) = p(1-p)$. During stochastic gradient descent, this high variance leads to unstable updates. To suppress this noise, we sample $N$ independent historical references $\{Y'_1, Y'_2, \dots, Y'_N\}$ to construct a multi-sample estimator $\bar{B} = \frac{1}{N} \sum_{i=1}^{N} \mathbb{I}(y > Y'_i)$. Given the i.i.d. nature of the samples, the variance of the averaged estimator is:
\begin{equation}
\text{Var}(\bar{B}) = \frac{\text{Var}(B)}{N} = \frac{p(1-p)}{N}
\end{equation}
By setting $N=50$ in our production system, we achieve a 50-fold reduction in estimation variance. This reduction provides a significantly smoother gradient surface, enabling the model to learn fine-grained ranking signals from noisy streaming data.

\textbf{Multi-sample Binary Cross-Entropy Loss.} 
To optimize the model using $N$ samples, we define the Multi-sample Binary Cross-Entropy (MBCE) loss as the expectation over the contrastive pool:
\begin{equation}
    \mathcal{L} = \frac{1}{N} \sum_{i=1}^N \left[ - B_i \log(\hat{p}) - (1 - B_i) \log(1 - \hat{p}) \right]
\end{equation}
where $B_i = \mathbb{I}(y > Y'_i)$. By the linearity of the BCE objective with respect to the label, this is mathematically equivalent to a single BCE loss with a soft label $\bar{p} = \frac{1}{N} \sum_{i=1}^N B_i$:
\begin{equation}
    \label{eq:mbce}
    \mathcal{L} = - \bar{p} \log(\hat{p}) - (1 - \bar{p}) \log(1 - \hat{p})
\end{equation}
According to the first-order optimality condition $\frac{d\mathcal{L}}{d\hat{p}} = 0$, the unique global optimum remains $\hat{p}^* = \bar{p}$, which is the empirical percentile of the current contrastive pool. Thus, MBCE stabilizes training and consistently converges to the desired percentile level.

\subsection{Value-Weighted Contrastive Learning}
\label{sec:Value-Weighted}
While the label-only contrastive approach effectively mitigates user-specific behavioral biases via relative rankings, it is inherently scale-agnostic, treating all interactions within the same percentile as having identical importance, regardless of their absolute magnitude differences. In many business-driven recommender systems, however, the absolute magnitude of a target (e.g., watch time) provides a complementary perspective on the economic value of an interaction. To bridge this gap, we generalize the framework into a Value-Weighted extension. This approach aims to find a balance between the bias-free nature of percentiles and the absolute magnitudes, ensuring that while the model still estimates a relative position, it also accounts for "how much" value was contributed by each interaction.

\textbf{Value-Weighted Binary Cross-Entropy Loss.} 
To capture the cumulative value contribution, we define the Value-Weighted Binary Cross-Entropy \textbf{(VWBCE)} loss by assigning the magnitude of each historical reference $Y'_i$ as a specific weight to its corresponding contrastive pair:
\begin{equation}
\begin{aligned}
    \mathcal{L}_{vw} = \frac{1}{\sum_{j=1}^N Y'_j} \sum_{i=1}^N Y'_i \Big[ &- \mathbb{I}(y > Y'_i) \log(\hat{p}) \\
    &- (1 - \mathbb{I}(y > Y'_i)) \log(1 - \hat{p}) \Big]
\end{aligned}
\end{equation}
By the linearity of the BCE objective, this is mathematically equivalent to a single BCE loss with a \textbf{value-weighted empirical label} $\bar{p}_{vw}$:
\begin{equation}
    \mathcal{L}_{vw} = - \bar{p}_{vw} \log(\hat{p}) - (1 - \bar{p}_{vw}) \log(1 - \hat{p})
\end{equation}
where the value-weighted empirical label $\bar{p}_{vw}$ is defined as $(\sum_{i=1}^N Y'_i \cdot \mathbb{I}(y > Y'_i)) / (\sum_{i=1}^N Y'_i)$.
This formulation ensures that the model's optimum $\hat{p}^*$ no longer represents the simple count-based percentile, but rather the proportion of total value contributed by interactions with magnitudes less than $y$.

\textbf{Theorem 3 (Global Optimum as Partial Expectation Ratio).} 
To understand how VWBCE captures value, we analyze its global optimum. We prove that the unique optimal prediction $\hat{p}^*$ transcends the standard CDF and corresponds to the Partial Expectation Ratio:
\begin{equation}
    \hat{p}^* = \frac{\int_{0}^{y} t f_u(t) dt}{\int_{0}^{\infty} t f_u(t) dt}
\end{equation}

\textit{Proof (Sketch):} By minimizing the total expected loss $\mathbb{E} [\mathcal{L}_{vw}]$, the first-order optimality condition for $\hat{p}$ yields the ratio of the partial expectation to the global expectation. This ratio represents the cumulative proportion of the total magnitude (e.g., watch time) contributed by all instances with magnitudes no greater than $y$. Consequently, the model unifies ordinal ranking and absolute magnitude contribution into a single objective, providing a cohesive metric for joint engagement and value optimization.

\subsection{Bootstrapped Contrastive Learning}
\label{sec:Bootstrapped Contrastive}
The contrastive paradigms discussed in previous sections assume a relatively continuous target distribution. However, in industrial recommender systems, many critical interaction signals—such as gifting, shares, or negative feedback—are highly sparse and discrete (often binary 0/1). For such classification-oriented targets, standard percentile estimation faces two fundamental challenges:

\begin{itemize}[leftmargin=*]
    \item \textbf{Loss of Ranking Resolution:} For a binary signal, the contrastive indicator $\mathbb{I}(y > Y')$ becomes a degenerate step function. A negative interaction ($y=0$) is always mapped to the 0th percentile, while a positive one ($y=1$) is always the 100th percentile. This binary "all-or-nothing" supervision fails to provide any fine-grained ranking information to distinguish preferences among users with the same binary label.
    \item \textbf{Empirical Distribution Degeneracy:} Due to label sparsity, a finite historical queue ($N=50$) for rare events often contains exclusively negative samples (e.g., 50 zeros). This results in a constant contrastive target of zero, leading to vanishing or unstable gradient updates and preventing the model from learning meaningful ranking boundaries.
\end{itemize}

To bridge the gap between discrete interactions and continuous preference ranking, we propose a prediction-based bootstrapping mechanism. Instead of contrasting raw binary labels, we leverage a continuous prior prediction to redefine the percentile.

\textbf{Softened Contrastive Target.} 
Let $\Psi(\cdot)$ be a prior regression model that predicts the continuous expected magnitude $\hat{y}$ for the current instance. To define the bootstrapped indicator, we contrast $\hat{y}$ against the prior predictions of historical samples $\hat{Y}' \sim f_{\Psi}$. For a single pair, the indicator is $\mathbb{I}(\hat{y} > \hat{Y}')$. To suppress estimation noise, the \textbf{multi-sample variance reduction} technique introduced in Section 3.3 can be similarly applied here:
\begin{equation}
    \bar{p}_{boot} = \frac{1}{N} \sum_{i=1}^{N} \mathbb{I}(\hat{y} > \hat{Y}'_i)
\end{equation}
where $\hat{Y}'_i$ are the continuous predicted values of the historical samples in the queue. By calculating the percentile within the prediction space rather than the raw label space, we transform discrete binary comparisons into a continuous and differentiable probability density.

\textbf{Informative Supervision.} 
This bootstrapping approach provides a much smoother supervisor. Even when ground-truth labels are identical (e.g., multiple samples with zero rewards), their prior predictions $\hat{y}$ can still reflect fine-grained differences in potential interest. By optimizing the model to fit these "softened" percentiles, we ensure the ranking head receives informative signals even in the most sparse feedback scenarios. Note that $\bar{p}_{boot}$ replaces $\bar{p}$ in the MBCE loss (\autoref{eq:mbce}) specifically for discrete targets.

\subsection{Percentile Co-Training}
\label{sec:Multi-Task Contrastive}
While the PEARL framework effectively mitigates user-specific behavioral biases through relative ranking, predicting the absolute magnitude of user engagement remains indispensable in many industrial ranking systems. This necessity is primarily driven by business logic where the predicted value directly translates to economic utility or monetary revenue.

\textbf{Motivation: Why Absolute Magnitudes Matter.} 
In many scenarios, both ordinal ranking and absolute magnitude are highly crucial.
A primary example is an advertising pricing system based on the expected Cost Per Mille (eCPM) framework. In such systems, the ranking score is typically the product of a predicted interaction rate and the absolute bid magnitude. While a scale-agnostic model can identify which ad is more likely to be clicked, it cannot provide the calibrated dollar-value estimation required for auction settlement and revenue maximization. Similarly, in livestream gifting or e-commerce, an interaction representing a \$100 transaction is fundamentally more valuable than one representing \$1, even if they share the same relative rank in different users' histories. Therefore, maintaining absolute scale calibration is essential for value-driven recommendation tasks.

\textbf{Joint Optimization via Co-training.} To leverage the strengths of both paradigms, we propose a co-training strategy that integrates PEARL as a debiased ranking regularizer alongside existing regression objectives. The joint loss function is defined as:
\begin{equation}
    \mathcal{L}_{\text{total}} = \mathcal{L}_{\text{reg}}(\hat{y}, y) + \lambda \cdot \mathcal{L}_{\text{PEARL}}(\hat{p}, y, \mathcal{P}_u)
\end{equation}
where:
\begin{itemize}[leftmargin=*]
    \item $\mathcal{L}_{\text{reg}}(\hat{y}, y)$ ensures the model retains its calibration on absolute scales by minimizing the error between the predicted magnitude $\hat{y} = f_{\text{reg}}(\mathbf{x}; \theta)$ and the ground-truth $y$.
    \item $\mathcal{L}_{\text{PEARL}}(\hat{p}, y, \mathcal{P}_u)$ acts as a ranking regularizer, optimizing the predicted percentile $\hat{p} = f_{\text{pct}}(\mathbf{x}; \theta)$ based on the current label $y$ and the user's historical contrastive pool $\mathcal{P}_u = \{Y'_i\}_{i=1}^N$.
\end{itemize}

At inference time, while the model benefits from the debiased representations learned through dual-head training, we primarily serve the predicted absolute magnitudes $\hat{y}$. 

\textbf{Synergistic Benefits.} The multi-task formulation offers complementary advantages by decoupling relative preference learning from absolute scale calibration:
\begin{itemize}[leftmargin=*]
    \item \textbf{User-level Ordinal Consistency:} By forcing the shared backbone to capture relative preferences within each user's specific distribution, PEARL prevents the model from being dominated by the gradients of high-activity users. This significantly improves the \textit{intra-user} ranking stability.
    \item \textbf{Scale Calibration:} The regression objective anchors the model's output to the physical domain, ensuring that the final score remains a calibrated magnitude required for accurate revenue estimation and auction settlement.
\end{itemize}

\subsection{Billion-Scale System Implementation}
\label{sec:Billion-Scale System}
Deploying the PEARL framework in a recommender systems with a combined user base of billions and strict millisecond-level latency requirements necessitates a minimalist yet robust engineering architecture. Our implementation focuses on two critical challenges in large-scale streaming: efficient state management and adaptive gradient gating.

\textbf{In-graph Storage via Reservoir Sampling.} 
To ensure mathematical consistency with our i.i.d. sampling assumption (Theorem 1) while adapting to evolving user interests, we implement the contrastive pools within the computational graph using Reservoir Sampling. For each user, the Parameter Server (PS) stores two user-specific states within the computational graph: (1) a contrastive pool $\mathcal{P}_u$ of size $N=50$, which stores the true magnitudes of the user's historical engagements, and (2) a counter $M_u$ that tracks the total number of streaming interactions processed for user $u$. 

Upon the arrival of the current interaction with magnitude $y$, the user-specific state is updated as follows:
\begin{equation}
    M_u \leftarrow M_u + 1
\end{equation}
If $M_u \le N$, the current magnitude $y$ is sequentially appended directly into the contrastive pool $\mathcal{P}_u$. For subsequent streaming instances where $M_u > N$, the active interaction $y$ enters the pool with a gating probability of $P = N / M_u$. If accepted, it overwrites a randomly selected slot within $\mathcal{P}_u$; otherwise, it is discarded. This classical in-graph streaming execution guarantees that at any timestep, $\mathcal{P}_u$ consistently maintains a uniform random sample of the user's entire historical distribution with near-zero retrieval latency and minimal memory footprint.

\textbf{Gradient Gating for Training Stability.} 
The statistical reliability of the empirical percentile $\bar{p}$ is inherently sensitive to the sample density within the reservoir. To mitigate high-variance updates from sparse historical states, we implement an adaptive gradient gating strategy. Based on our analysis of estimation stability, we define a triggering threshold $\tau = 10$. A training instance contributes to the gradient update of the ranking head if and only if its contrastive pool $\mathcal{P}_u$ has attained sufficient statistical support (i.e., $M_u \ge \tau$). For instances failing to meet this threshold, the system entirely bypasses the ranking-head optimization for that instance. This selective update mechanism prevents noisy empirical signals from under-sampled users from polluting the shared backbone, ensuring stable convergence in billion-scale streaming environments.
\section{Evaluation}

In this section, we evaluate our proposed PEARL framework against other popular debiasing techniques on multiple relevant targets for online livestream recommendation. All our experiments are conducted in our industry recommender system---a leading livestream platform that attracts billions of users globally---to demonstrate its real-world effectiveness. 

\subsection{Evaluation Setup}
\subsubsection{Dataset}
For training and evaluation, we use a production streaming dataset from the ranking stage of our industrial recommender system. Statistics are provided in \autoref{tab:dataset}.

\begin{table}[h]
\caption{Statistics of the dataset. } \label{tab:dataset}
\centering
\begin{tabular}{cccc}
\toprule
 & \textbf{\# Instances} & \textbf{\# Features}  & \textbf{\# Targets}\\
\midrule
Industry & 168B & 4552 & 98 \\
\bottomrule
\end{tabular}
\end{table}

\subsubsection{Metrics}
To assess each method's performance relative to their original target, we evaluate across multiple metrics:
\begin{itemize}[leftmargin=*]
    \item \textbf{UAUC:} User-Averaged AUC, averaging per-user AUC scores across the evaluation set. Unlike AUC, UAUC accounts for variability in user behavior and interaction history, providing a more faithful measure of personalized ranking quality. For regression targets, we instead calculate User-Averaged Regression AUC (URegAUC) to make comparison between targets more fair. Following the definition of \cite{zhan2022deconfounding}, we define Regression AUC as the fraction of uniformly sampled item pairs for which the predicted watch-time values preserve the ground-truth ranking order.
\end{itemize}

All results are reported as both raw values and relative improvements over our baseline model in production.

\subsubsection{Baselines}
We compare against the original regression or classification baselines of the targets, as well as other strong debiasing techniques proposed for content recommendation. Specifically, we include the following baselines: the original regression model, Conditional Quantile Estimation (CQE) ~\cite{lin2024conditional}, and Relative Advantage Debiasing (RAD) ~\cite{rad2025}. All the methods are based on a modern ranking architecture backbone \cite{zhang2026zenith} as the shared bottom, with individual task-specific Multilayer Perceptron (MLP) towers to adapt to different targets.

\subsubsection{Training}
For all methods, we use the RMSPropV2 optimizer with a learning rate of 0.0015, momentum of 0.999999, and an initialization factor of 1.0. 

\begin{table}[h]
\caption{Performance comparison for the watch time target between PEARL and baselines.} 
\label{tab:method_ablation}
\centering
\begin{tabular}{l c}
\toprule
\textbf{Method} & \textbf{UAUC} \\
\midrule
Original & 0.641 \\
CQE & 0.615 \\
RAD & 0.625 \\
PEARL (ours) & 0.648 \\
\bottomrule
\end{tabular}
\end{table}

\begin{table*}[h]
\caption{PEARL's UAUC improvements across multiple targets, both for binary classification and regressive predictions. For the co-training variation, we list the performance for the original target with co-training.} 
\label{tab:target_ablation}
\centering
\begin{tabular}{c c c c c c}
\toprule
\textbf{Target} & \textbf{PEARL Variation} & \textbf{Baseline UAUC} & \textbf{PEARL UAUC} \\
\midrule
Interaction & Multi-sample Percentile (Sec 3.3) & 0.606 & 0.648 (+6.93\%) \\
Session Watch Time & Value-Weighted Percentile (Sec 3.4) & 0.641 & 0.644 (+0.468\%) &  \\
Consumption & Value-Weighted Percentile (Sec 3.4) & 0.583 & 0.589 (+0.977\%) \\
Report & Prediction-Based Bootstrapping (Sec 3.5) & 0.743 & 0.780 (+4.96\%) \\
Long-term Watch Time & Percentile Co-Training (Sec 3.6) & 0.689 & 0.800 (+16.1\%) \\
\bottomrule
\end{tabular}
\end{table*}

\begin{table*}[h]
\caption{Online A/B performance gains across multiple targets and corresponding metrics. } 
\label{tab:online_metrics}
\centering
\begin{tabular}{c c c c c c c c c c}
\toprule
\textbf{Target} & \textbf{Watch Session} & \textbf{Watch Duration} & \textbf{Consumption Amount} & \textbf{Interaction Rate} & \textbf{Report Rate}\\
\midrule
Session Watch Time & +0.608\% & +0.665\% & --- & --- & ---\\
Long-term Watch Time & +1.165\% & +1.222\% & --- & --- & --- \\
Consumption & +0.098\% & +0.232\% & +0.796\% & --- & --- \\
Interaction & --- & --- & --- & +1.494\% & --- \\
Report & --- & --- & --- & --- & -6.911\% \\
\bottomrule
\end{tabular}
\end{table*}

\subsection{Offline Results}
As shown in \autoref{tab:method_ablation}, our debiasing method achieves the highest UAUC for the session watch time target, one of the most important target in our online ranking system. This reflects PEARL's ability to address the watch time imbalance between low-active and high-active livestream users by personalizing their predictions to their history. In comparison, the other ablation methods underperform for various reasons. While CQE is able to debias many features at once, the non-personalized nature of the debiasing makes it underperform for low-active livestream users and overperform for high-active livestream users, as reflected in its low UAUC. In the context of livestream recommendation, RAD is unable to take full advantage of its user-sided CDF due to the lack of watch-time bins. 

\autoref{tab:target_ablation} further demonstrates the effectiveness of PEARL and its variations in other important targets. We applied different PEARL variations to multiple targets based on the complexity of the label distribution. Specifically, for the interaction target, we chose the multi-sample percentile variation, where the gain from this variation is significant enough that further adding value-weighting would introduce complexity with diminishing returns. For the session watch time and consumption targets, as the label distribution is more difficult and complex to model, we chose to use the value-weighted percentile variation to increase the target performance as much as possible. As a binary target, report was implemented using the prediction-based bootstrapping variation to transform the binary label into a continuous label. For the long-term watch time target, we utilized the Percentile Co-Training variant as a priority was to keep its calibration stable. 

Overall, PEARL consistently outperforms the existing regression targets for all the targets. We see the largest performance increase from the long-term engagement and report targets, which benefit from user personalization the most due to their sparse label distributions. PEARL focuses on per-user livestream preferences for these targets, allowing the model to contextualize each user's watch duration relative to their historical behavior rather than the global distribution. This alleviates the label imbalance introduced by the uneven distribution of engagement level across users, as the model no longer needs to reconcile sparse positive labels from low-active users against the dominant signal from high-active users. Contrastingly, session watch time and consumption targets show smaller improvements. While consumption labels are globally sparse, its positive labels are concentrated within a specific user subgroup, creating a high-density distribution within that group and reducing the benefits of personalization. For the long-term watch time target, the actual UAUC for the regression target is on par with the co-trained percentile target, demonstrating the strong knowledge transfer from percentile modeling for alleviating user bias. 

\begin{figure}[t]
  \centering
    \includegraphics[width=0.95\linewidth]{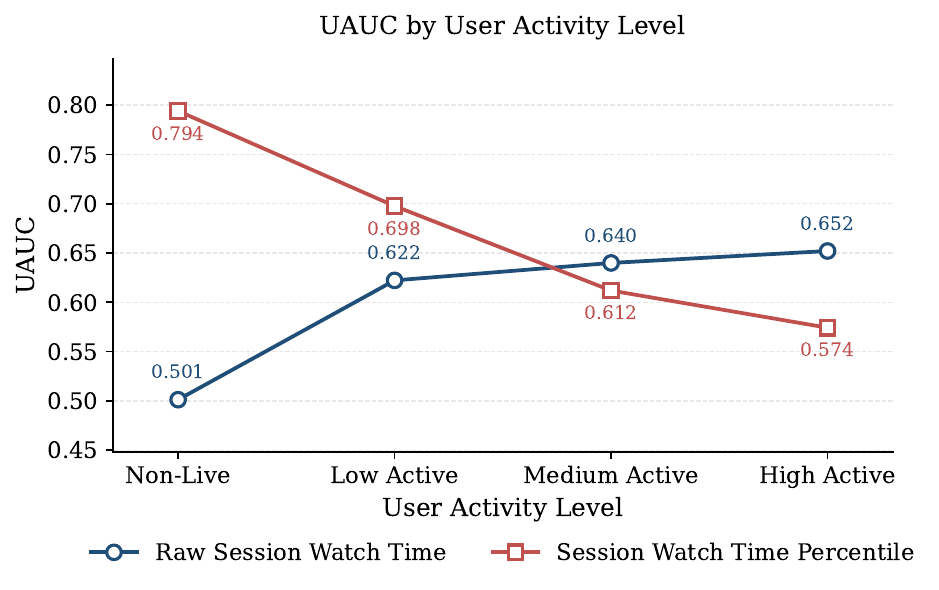}
    \caption{The comparison of the UAUC among different user active level cohorts. The non-live cohort corresponds to users who have not watched livestreams in the past month, while the high-active cohort corresponds to users who consumes livestreams content on a daily basis.}
    \label{fig:uregauc_by_activity}
\end{figure}

To further demonstrate the practical effectiveness of PEARL in addressing the user bias in modeling, we show the UAUC comparison between the percentile target and the original regression target for session watch time across different user active level cohorts in \autoref{fig:uregauc_by_activity}. Notably, the raw watch time target suffers from strong user bias, where model learning is dominated by the magnitudes of interaction signals. This leads to better performance for users with stronger engagements, but ranking capability for non-live users reduces to random guessing. On the other hand, the corresponding percentile target by PEARL shows clear strengths for non-live and low-active users (the majority of our user base), providing a valuable complement to existing value-dominated regression targets. 

\subsection{Online A/B Results}
We evaluate the performance of PEARL in real-world scenarios by conducting multiple online week-long A/B tests that complement or replace the baseline target predictions with our debiased predictions in our production system. As PEARL is applied to different targets, we present their corresponding online metrics as follows: Watch Session/User, Watch Duration/User, Consumption Amount/User, Interaction Rate/User, and Report Rate/User. The online results are presented in \autoref{tab:online_metrics}. Overall, PEARL achieves significant online metric gains in all tested categories, demonstrating its strengths in improving real-world user engagement by reducing target biases in modeling.
\section{Conclusion}

In this work, we proposed a nonparametric contrastive percentile modeling framework that models relative preference signals through pairwise comparisons over real interaction samples, capturing percentile relationships without explicit engagement distribution estimation. We further provided theoretical justification for the consistency of the proposed formulation. For broader applicability, we introduced several extensions, including a generalized value-weighted formulation to incorporate heterogeneous interaction importance, a prediction-based bootstrapping mechanism for percentile smoothing under sparse and discrete feedback, and a co-training strategy to improve representation learning, along with industry system incorporation considerations. Extensive experiments demonstrate that our approach consistently improves recommendation performance and robustness, highlighting the effectiveness of contrastive percentile learning for debiasing user behavior.

\bibliographystyle{ACM-Reference-Format}
\bibliography{ref}

\appendix
% Empty









\end{document}